# Improving Model Drift for Robust Object Tracking

Qiujie Dong, Xuedong He, Haiyan Ge, Qin Liu, Aifu Han and Shengzong Zhou

*Abstract*—Discriminative correlation filters show excellent performance in object tracking. However, in complex scenes, the apparent characteristics of the tracked target are variable, which makes it easy to pollute the model and cause the model drift. In this paper, considering that the secondary peak has a greater impact on the model update, we propose a method for detecting the primary and secondary peaks of the response map. Secondly, a novel confidence function which uses the adaptive update discriminant mechanism is proposed, which yield good robustness. Thirdly, we propose a robust tracker with correlation filters, which uses hand-crafted features and can improve model drift in complex scenes. Finally, in order to cope with the current trackers' multi-feature response merge, we propose a simple exponential adaptive merge approach. Extensive experiments are performed on OTB2013, OTB100 and TC128 datasets. Our approach performs superiorly against several state-of-the-art trackers while runs at speed in real time.

## I. INTRODUCTION

WITH the advent of the automation era, object tracking has gradually become a hot topic in recent years [1]-[11]. Tracking is multidisciplinary research that predicts a target position in all subsequent frames given the initial frame information. In the early 21st century, the tracking models are generative models, and their performances are generally poor. Inspired by object detection, the discriminative models that separate the target from the background have become the mainstream.

This work was supported by the Chinese Academy of Sciences STS Projects 2019T31020008, 2019T31020010.

Corresponding author: Shengzong Zhou.

Q. Dong, X. He, Q. Liu, A Han and S. Zhou with the Fujian Institute of Research on the Structure of Matter Chinese Academy of Sciences, Fuzhou 350002, China (e-mail: dongqj@fjirsm.ac.cn, hexd@fjirsm.ac.cn, liuq@fjirsm.ac.cn, hanaf@fjirsm.ac.cn, zhousz@fjirsm.ac.cn).

Q. Dong, X. He and A Han with the North University of China, Taiyuan 030051, China.

H. Ge with the Shandong University of Technology, Zibo 255049, China (e-mail: 1178163538@qq.com)

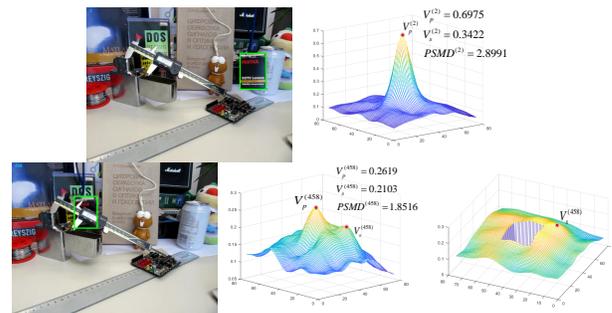

Fig.1 Response map of different scenes. (a)tracking result (b)response map for frame 2 of box sequence, (c)tracking result (d)response map (e) Secondary Peak response map for frame 458 of box sequence.

At present, mainstream trackers based on discriminative models include the correlation filters (CF) trackers [12]-[22] which use hand-crafted features or CNNs and the Siamese Network trackers [23]-[28]. The CF trackers can update the tracking models in real time, so their robustness is better. The Siamese Network trackers adopt the offline pre-training networks, so their accuracy is higher, but the deep networks are larger, tracking models are hard to real-time updated, trackers' robustness is worse. The Visual Tracker Benchmark[1] includes 11 attributes.

Though the CF trackers are nearing maturity, the constant variation of the target's scenes cause easily the background information to be learned into the model, resulting in the models to be polluted [29]. The reason for the contaminated model is that the response map is no longer a single peak, but a multi-peak in the complex scenes. In the case where the difference between the value of primary peak and the value of secondary peak is not obvious, the model drift is most likely to occur.

Currently, the approach to solving model drift problems is to selectively update the model. Bolme et al. [12] proposed Peak to Sidelobe Ratio (PSR) confidence function, but the actual application effect is not prominent, so it has not been widely used. Wang et al. [21] proposed a novel criterion called average peak-to-correlation energy (APCE), which is better than PSR. Danelljan et al. [14] proposed a simple lazy update mechanism to update the model every five frames. However, model drift will still occur for [14] proposed scheme when the update frame is the image whose target apparent features change greatly.

Now the CF trackers utilize multiple features instead of a single feature. However, a fixed merge factor is widely used in feature fusion, which reduces the performance of the trackers in the complex tracking scenarios and worst of all causes tracking failure.

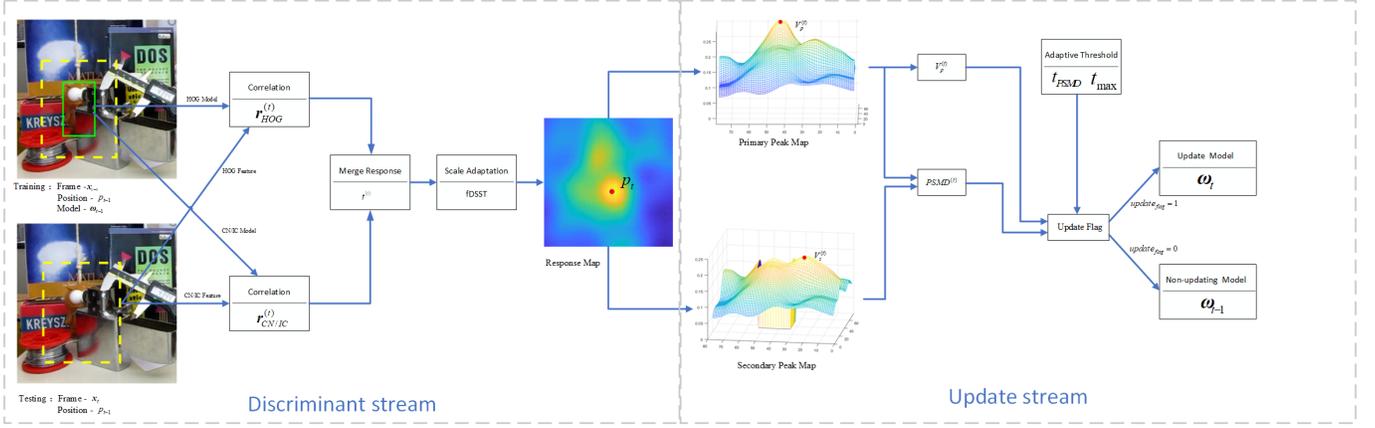

Fig.2 The overall architecture of the proposed model. The whole model contains two streams: Discriminant stream and update stream.

In this paper, we consider the problems mentioned above and propose a novel object tracking method called improving model drift for robust object tracking with correlation filters (MDRCF). The main contributions of our works can be summarized as follows:

- A novel feature response map peak detection method is proposed to realize the detection of primary and secondary peaks.
- We propose a new confidence function using the adaptive update discriminative mechanism that can select different thresholds according to different tracking videos. This confidence function has good robustness and can be well transplanted in the others CF trackers.
- Based on the above method, we explore a novel CF tracker (MDRCF) with the hand-crafted features including Histogram of Oriented Gradient (HOG) and Color Name (CN), which can effectively improve the model drift and enhance tracker's performances in complex scenes. The proposed algorithm is compared with state-of-the-art (SOTA) CF trackers on the large-scale datasets: OTB2013 [30], OTB100 [31] and TC128 [32]. The proposed tracker's performances exceed ECO-HC on all three datasets.
- Finally, aiming at the multi-feature merge problem, we establish a new and simple multi-feature response adaptive merge method which is based on Staple [13], which is called exponential adaptive merge Staple (EAMStaple). The experiment proves that our proposed tracker works well in solving the multi-feature merge problem.

## II. RELATED WORK

The CF trackers extract the template $\boldsymbol{\beta}$ from the initial frame and continuously train and update $\boldsymbol{\beta}$ in subsequent frames. In the $t^{th}$ frame, the template $\boldsymbol{\beta} \in \wp$ is updated where the loss function $L[g(\boldsymbol{x}_t, \boldsymbol{\beta}), \boldsymbol{y}_t]$ is minimized for the input sample $\boldsymbol{x}_t$

and the desired output $\boldsymbol{y}_t$, that is

$$\boldsymbol{\beta}_t = \arg\min_{\boldsymbol{\beta} \in \wp} \sum_{i=1}^{t} \{L[g(\boldsymbol{x}_t, \boldsymbol{\beta}), \boldsymbol{y}_t] + \lambda \|\boldsymbol{\beta}\|^2\} \quad (1)$$

where $\lambda$ is a regularization parameter to prevent over-fitting, and $\|\cdot\|$ is $L_2$ norm. $g(\boldsymbol{x}_t, \boldsymbol{\beta}) = \boldsymbol{\beta}^T \boldsymbol{x}_t$ is the regression function.

For the loss function, the ridge regression method is used in this paper, and its performance is consistent with the Support Vector Machine (SVM), but the structure is relatively simple and has a closed-form solution [33],

$$L[g(\boldsymbol{x}_t, \boldsymbol{\beta}), \boldsymbol{y}_t] = [\boldsymbol{\beta}^T \boldsymbol{x}_t - \boldsymbol{y}_t]^2 \quad (2)$$

Substituting Eq. (2) into Eq. (1),

$$\boldsymbol{\beta}_t = \arg\min_{\boldsymbol{\beta} \in \wp} \sum_{i=1}^{t} [(\boldsymbol{\beta}^T \boldsymbol{x}_t - \boldsymbol{y}_t)^2 + \lambda \|\boldsymbol{\beta}\|^2] \quad (3)$$

According to the [33], the closed-form solution of Eq. (3) is

$$\boldsymbol{\beta} = (\boldsymbol{x}^T \boldsymbol{x} + \lambda \boldsymbol{I})^{-1} \boldsymbol{x}^T \boldsymbol{y} \quad (4)$$

Where $\boldsymbol{I}$ is an identity matrix of the same dimension as $\boldsymbol{x}^T \boldsymbol{x}$.

Mapping $\boldsymbol{\beta}$ to nonlinear space, the variable to be solved is transformed from $\boldsymbol{\beta}$ to $\boldsymbol{\omega}$, and according to the kernel technique mentioned in [33], Eq. (4) is

$$\boldsymbol{\omega} = (\boldsymbol{K} + \lambda \boldsymbol{I})^{-1} \boldsymbol{y} \quad (5)$$

Where $\boldsymbol{K}$ is kernel matrix.

The kernel matrix $\boldsymbol{K}$ is a symmetric matrix, which can be diagonalized. The Eq. (5) by the Discrete Fourier Transform (DFT) can implement time domain convolution to frequency domain Hadamard product, reducing the computational complexity by a margin. Then Eq. (5) expresses

$$\hat{\boldsymbol{\omega}} = \frac{\hat{\boldsymbol{y}}}{\hat{\boldsymbol{K}} + \lambda} \quad (6)$$

where hat $^\wedge$ represents the corresponding DFT vector of the matrix and $\hat{\boldsymbol{K}}$ is the Gaussian kernel matrix.

In order to preserve the information of the previous frames, the model is updated using a linear interpolation method,

$$\hat{\boldsymbol{\omega}}_t = \begin{cases} \hat{\boldsymbol{\omega}} & , t = 1 \\ (1-\eta)\hat{\boldsymbol{\omega}}_{t-1} + \eta\hat{\boldsymbol{\omega}}, t > 1 \end{cases} \quad (7)$$

Where $\hat{\boldsymbol{\omega}}_t$ is the current frame template and $\hat{\boldsymbol{\omega}}_{t-1}$ is the previous frame template. $\eta \in [0,1]$ is the learning rate.

The CF tracker's time domain response map of the current

frame is

$$r_t = \mathcal{F}^{-1}(\hat{\omega}_t \odot \hat{K}) \tag{8}$$

Where $\odot$ is the Hadamard product.

The detailed derivation formulas can be found in [19], [20], [33].

## III. OUR APPROACH

In this section, we first elaborate the proposed feature response map peak detection method and demonstrate its effect. Next, we deduce a novel confidence function which used to achieve the selective update of the model. Thirdly, the adaptive update discriminating mechanism is proposed to break the tradition of discriminating using a fixed discriminant threshold. In the end, we present a new and simple target multi-feature adaptive merge method.

### A. Primary and Secondary Peaks Detection

In the ideal case, the CF tracker obtained by Eq. (8) has only one peak in the time domain response, as visualized in Figure 1(b). However, in the complex scenes of OCC, SV, IV and etc., the time domain response maps are introduced unwanted multi-peak (see Figure 1(d)).

In the $(t+1)^{th}$ frame, the input sample $x_{t+1}$ is traversed using a cyclic matrix, instead of a sliding window, to generate an image segment data set $S_{t+1}$, and the image segment $s$ in $S_{t+1}$ is matched with the model $\beta_t$ so that the matching value function $f(s, \beta_t)$ is maximized, that is

$$s_{t+1} = \arg \max_{s \in S_{t+1}} f(s, \beta_t) \tag{9}$$

However, when there exists a secondary peak and its value is larger, it is updated into the model $\beta_t$, and when the prediction of the $(t+1)^{th}$ frame is performed, the image segment $s^*_{t+1}$ may be matched,

$$s^*_{t+1} = \arg \max_{s \in S_{t+1}} f(s, \beta_t) \tag{10}$$

Compared with $s_{t+1}$, there is

$$f(s^*_{t+1}, \beta_t) > f(s_{t+1}, \beta_t) \tag{11}$$

This will result in a tracking offset, and the worst is that the target will be lost. Therefore, it is important to perform primary and secondary peaks detection on the time domain response of the CF tracker.

For the time domain response $r_t$, the detection of the primary peak's position is relatively easy. Extract the maximum response layer in $r_t$,

$$l = \arg \max_{w \in W, h \in H, l \in L} r_t(w, h, l) \tag{12}$$

Where $W$, $H$ and $L$ are the width, height and the number of layers of $r_t$, respectively.

Then $r_l^{(t)} \in r_t$ is the maximum response layer of $r_t$. The time domain response map's primary peak is located in the domain response maximum response layer,

$$[l_{lw}^{(t)}, l_{lh}^{(t)}] = \arg \max_{lw \in W, lh \in H} [r_l^{(t)}(lw, lh)] \tag{13}$$

where $[l_{lw}^{(t)}, l_{lh}^{(t)}]$ is the primary peak's position of $r_t$.

Different from the [21], the mask matrix $\psi$ is a binary matrix with the same dimension as $r_l^{(t)}$ of the time domain response $r_t$. The elements at the location of the rectangular region centered on $[l_{lw}^{(t)}, l_{lh}^{(t)}]$ and having the length of the $2N \times 2N$ side are set to 0, while others are set to 1. $N$ is a hyperparameter with a value range of $[2, 15]$.

The secondary peak's response map matrix is the Hadamard product of the maximum response layer $r_l^{(t)}$ and the mask matrix $\Psi$,

$$r_s^{(t)} = r_l^{(t)} \odot \Psi \tag{14}$$

Here, $r_s^{(t)}$ is the secondary peak's response map matrix (see Figure 1(e)).

The time domain response map's secondary peak is located in $r_s^{(t)}$,

$$[l_{sw}^{(t)}, l_{sh}^{(t)}] = \arg \max_{sw \in W, sh \in H} [r_s^{(t)}(sw, sh)] \tag{15}$$

Here, $[l_{sw}^{(t)}, l_{sh}^{(t)}]$ is the secondary peak's position of $r_t$.

### B. Confidence Function

The traditional models update each frame. Although this method can continuously learn according to the target's appearance characteristics, it is also easy to cause pollution of the model. Reference [12] proposes to use the PSR for peak intensity detection, but the effect is not good. Reference [21] proposed the APCE confidence function, but this function is mainly concerned with the average value, while ignoring the most influence on the primary peak is the secondary peak of $r_t$.

In the $t^{th}$ frame, in order to have a unified evaluation criterion for the confidence function, we average the maximum response layer $r_l^{(t)}$ of $r_t$,

$$V_m^{(t)} = \frac{1}{W \cdot H} \sum_{lw=1, lh=1}^{W, H} [r_l^{(t)}(lw, lh)] \tag{16}$$

Where $V_m^{(t)}$ is the mean of the maximum response layer $r_l^{(t)}$ of $r_t$.

The maximum response value $V_p^{(t)}$ of $r_t$ is

$$V_p^{(t)} = \max_{w \in W, h \in H, l \in L} r_t(w, h, l) \tag{17}$$

In the maximum response layer $r_l^{(t)}$, the maximum response value $V_s^{(t)}$ other than $V_p^{(t)}$ is

$$V_s^{(t)} = \max_{w \in W, w \neq lw, h \in H, h \neq lh} r_l^{(t)}(w, h) \tag{18}$$

The new confidence function proposed in this paper is called Primary and Secondary Peak Mean Difference Ratio (PSMD), and its expression is

$$PSMD^{(t)} = \frac{V_p^{(t)} - V_m^{(t)}}{\left| V_s^{(t)} - V_m^{(t)} \right|} \tag{19}$$

Here, $|\cdot|$ is the symbol for absolute value.

Figure 1(b) and Figure 1(d) show the PSMD of the frames in different scenarios, respectively.

### C. Adaptive Update Discriminating Mechanism

Using the confidence function to perform model update discrimination, both [12] and [21] use a fixed discriminant threshold, which is not good for processing different tracking sequences. In this paper, we get adaptive update discrimination threshold according to different sequences, which is better robustness.

Since the initial frame in the tracking is manually specified, image processing is generally not performed on the initial frame. The tracking effect of the second frame is optimal. In this paper, we use the data of the second frame to find the adaptive update threshold.

The adaptive update discrimination thresholds $t_{PSMD}$ of the confidence function PSMD is

$$t_{PSMD} = (\frac{V_p^{(2)} - V_m^{(2)}}{\left| V_s^{(2)} - V_m^{(2)} \right|}) / \mu \quad (20)$$

Where $\mu$ is a hyperparameter with a range of $[1,11]$.

It is not enough to just update the model with a confidence function. Since the value of PSMD is still large in the early stages of complex scenes such as occlusion, if the model is still updated, it will gradually lead to the model to be polluted. In the case, the maximum response value will change significantly. Therefore, based on the PSMD, we increase the maximum response value adaptive update threshold $t_{max}$,

$$t_{max} = V_p^{(2)} / \nu \quad (21)$$

Where $\nu$ is a hyperparameter with a range of $(0,5]$.

The model judgment criteria $V_p^{(t)}$ and $PSMD^{(t)}$ are respectively compared with the adaptive update threshold to obtain the update flag $update_{flag}$, that is

$$\begin{cases} update_{flag} = 1, & if \quad PSMD^{(t)} \geq t_{PSMD} \ \& \quad V_p^{(t)} \geq t_{max} \\ update_{flag} = 0, & others \end{cases} \quad (22)$$

If $update_{flag} = 1$, we use linear interpolation to update the model, the current frame is updated to the model while retaining the previous frame. If $update_{flag} = 0$, the model is not updated.

### D. MDRCF

We propose a novel CF tracker called MDRCF. This algorithm uses only the hand-crafted features which are HOG and CN and doesn't use depth features.

It should be noted that when the tracking sequences are grayscale, using the CN will increase the unnecessary computational cost. In this paper, when the sequences are grayscale, the intensityChannelNorm6 (IC) [14] is used, which is a simplified CN.

In order to ensure the real-time performance of the algorithm, we employ the fast Discriminative Scale Space Tracking (fDSST) method proposed in [17] to solve the scale change problem in our tracker.

Hyperparameter $N$ in Section 3.1 and hyperparameters $\mu$ and $\nu$ in Section 3.2 can be calculated offline. We make use of

---

**Algorithm1:** Our MDRCF approach, iteration at time step t.

**Input:**

Frame $x_t$.

Previous target position $p_{t-1}$.

Model $\omega_{t-1}$.

**Output:**

Estimated target position $p_t$.

Updated model.

**Frame=1:**

Compute model $\omega_1$ with HOG and CN.

**Frame=2:**

Compute adaptive threshold $t_{PSMD}$ and $t_{max}$ using (20) and (21), respectively.

**Frame ≥ 2:**

1. Extract sample $s$ from $x_t$ at $p_{t-1}$.

2. Compute correlation scores $r_{HOG}^{(t)}$ and $r_{CN}^{(t)}$.

3. Compute merge response.

4. Compute the target scale using fDSST.

5. Set $P_t$ to the target position that maximizes $r_t$.

6. Compute $V_m^{(t)}, V_p^{(t)}$ and $V_s^{(t)}$ using (16), (17) and (18).

7. Compute confidence score $PSMD^{(t)}$ using (19).

8. Model update

8.1 If $PSMD^{(t)} \geq t_{PSMD} \ \& \ V_p^{(t)} \geq t_{max}$, set $\omega_t$ to the target model.

8.2 If $PSMD^{(t)} < t_{PSMD} \ | \ V_p^{(t)} < t_{max}$, set $\omega_{t-1}$ to the target model.

---

VOT2018 [34] as the training set and obtain three hyperparameters through training.

Figure 2 is the framework of our algorithm, and algorithm 1 is a brief overview of our algorithm.

### E. Multi-feature Adaptive Merge Method

Taking the Staple tracker as an example, it uses the target HOG feature and color histogram feature to perform feature response merge,

$$r = (1 - \tilde{\lambda})r_{cf} + \tilde{\lambda}r_{ch} \quad (23)$$

Where $r_{cf}$ is HOG response and $r_{ch}$ is color histogram response. $\tilde{\lambda}$ is the merge factor which is 0.3 in Staple.

In practice, the representation ability of HOG is better than the color histogram, so more attention needs to be paid to the HOG feature, it must be guaranteed that

$$(1 - \tilde{\lambda}) > \frac{1}{2} \quad (24)$$

When the target probability mean value $\alpha$ in the target color histogram feature is larger, it indicates that the target color feature and the background color feature have obvious differences at this time, so the attention to $r_{ch}$ should be appropriately increased. However, when background color interference occurs, $\alpha$ will also increase, but at this time, it is necessary to reduce the attention to $r_{ch}$ as much as possible to avoid updating the background information into the model.

Therefore, the merge factor $\tilde{\lambda}$ is solved using a piecewise function. To ensure the simplicity and smoothness of the model, an exponential adaptive merge factor is used,

$$\tilde{\lambda}^{(t)} = \begin{cases} \exp(\dfrac{1}{n_\tau}\sum_{\tau=1}^{n_\tau}\alpha_\tau^{(t)}) - \phi & , \alpha^{(t)} < \hbar \\ \exp(-\dfrac{1}{n_\tau}\sum_{\tau=1}^{n_\tau}\alpha_\tau^{(t)})/\varepsilon & , \alpha^{(t)} \geq \hbar \end{cases} \quad (25)$$

Where $\alpha_\tau^{(t)}$ is the probability value that pixel $\tau$ is the target in the $t^{th}$ frame, and $n_\tau$ is the number of pixels in the response matrix. $\hbar$ is the discriminant threshold and is a hyperparameter. $\phi$ and $\varepsilon$ are hyperparameters.

## IV. Experiments

In this section, the experimental parameters and evaluation criteria are given first. Furthermore, the confidence function proposed in this paper is compared with other confidence functions in the current tracking domain to evaluate the performance of the confidence function. Thirdly, the proposed algorithm (MDRCF) and SOTA CF algorithms are quantitatively and qualitatively compared and analyzed. Finally, we evaluate the performances of the EAMStaple proposed in this paper.

### A. Implementation Details

The software platform of this experiment is MATLAB R2017b. The hardware platform configuration is a desktop computer with Intel (R) Core (TM) i7-4790CPU@3.60GHz and 12GB RAM. The experimental hyperparameters of the algorithm are shown in Table I.

TABLE I
EXPERIMENTAL HYPERPARAMETERS

| Parameters | Values |
|---|---|
| $N$ | 10 |
| $\mu$ | 1.06 |
| $\nu$ | 0.94 |
| $\hbar$ | 0.38 |
| $\phi$ | 1.09 |
| $\varepsilon$ | 2 |

The evaluation datasets are OTB2013, OTB100 and TC128. The OTB2013 contains 51 sequences which contain 11 attributes. The OTB100 includes 100 sequences, which are more complex and difficult than the OTB2013. The TC128 has 128 color sequences which are rich in colors.

The experiments in this paper are based on the one-pass evaluation (OPE) protocol in [29] for quantitative analysis, including precision and area under the curve (AUC). In order to ensure fairness, the precision threshold and AUC threshold of the quantitative analysis are 20 pixels and 0.5 respectively, and all algorithms don't use the target's depth features.

### B. Comparison of Confidence Function

The current confidence function in object tracking are PSR proposed by Bolme et al. [12] and APCE proposed by Wang et al. [21]. Based on the Staple algorithm, the confidence function

PSMD proposed in this paper is compared with PSR and APCE. Figure 3 shows the precision and AUC values for the three confidence functions based on Staple on different datasets.

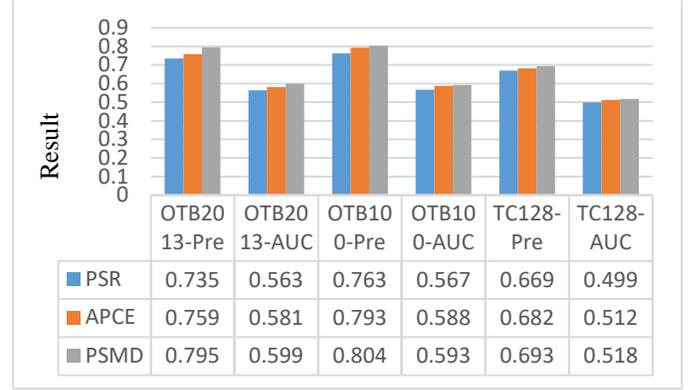

Fig.3 Tracking result of three confidence functions on different dataset.

| | OTB2013-Pre | OTB2013-AUC | OTB100-Pre | OTB100-AUC | TC128-Pre | TC128-AUC |
|---|---|---|---|---|---|---|
| PSR | 0.735 | 0.563 | 0.763 | 0.567 | 0.669 | 0.499 |
| APCE | 0.759 | 0.581 | 0.793 | 0.588 | 0.682 | 0.512 |
| PSMD | 0.795 | 0.599 | 0.804 | 0.593 | 0.693 | 0.518 |

It can be seen from Figure 3 that the PSMD confidence function is the best both the precision and the AUC evaluation criteria on three datasets compared with the PSR and APCE. It can be proved that the performance of the confidence function proposed in this paper is superior.

### C. Comparison of MDRCF and SOTA CF Trackers

In order to verify the performance of the proposed MDRCF, we select four SOTA CF trackers which are Staple, SRDCF, LMCF and ECO-HC to compare with our algorithm and perform quantitative and qualitative analysis.

Notes: Since the four SOTA algorithms we selected in the current CF trackers are excellent, the earlier CF trackers [12], [15], [19], [20], [35] are no longer considered. LMCF doesn't disclose the source code, so we use the experiment data in its paper for comparison.

First, quantitative analysis was performed on five trackers. According to the precision plots and AUC plots of the four trackers in OTB2013, OTB100 and TC128, the proposed algorithm has better performance than the SOTA trackers (see Figure 4).

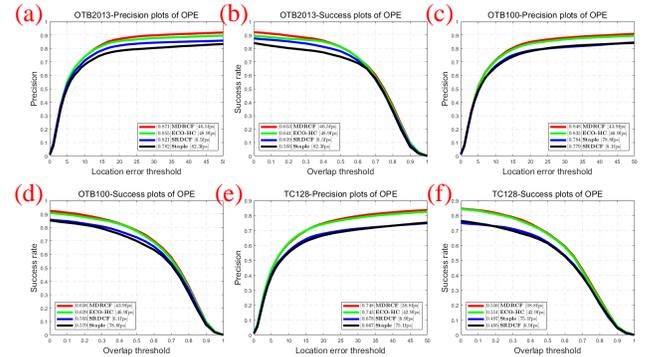

Fig.4 Tracking result of four trackers. (a)precision (b) AUC for OTB2013, (c) precision (d)AUC for OTB100, (e)precision(f)AUC for TC128

Table II also shows the precision values and AUC values of the five trackers in the three datasets and MDRCF performs better in the three datasets. Compared with ECO-HC, the precision values of MDRCF on the three datasets are 1.87%, 1.56% and 0.67% higher respectively, and the AUC values are 1.87%, 1.43% and 0.91% higher.

TABLE III
TRACKING RESULTS (IN %) OF FOUR TRACKERS STAPLE, SRDCF, ECO-HC AND MDRCF ON 11 ATTRIBUTES FOR OTB2013, OTB100 AND TC128 DATASETS. THE ENTRIES IN BOLD RED DENOTE THE BEST RESULTS AND THE ONES IN ITALIC BLUE INDICATE THE SECOND BEST.

| Trackers | Attributes | IV | OPR | SV | OCC | DEF | MB | FM | IPR | OV | BC | LR |
|---|---|---|---|---|---|---|---|---|---|---|---|---|
| Staple | OTB 2013 | 56.1 | 56.9 | 54.5 | 58.5 | 60.7 | 52.6 | 50.1 | *57.6* | 51.8 | 55.7 | *39.6* |
| SRDCF | | 56.3 | 59 | 57.3 | 61.5 | *63.7* | *56.8* | 54.6 | 55.3 | 55.5 | 57.1 | **47.1** |
| ECO-HC | | *58.9* | *61.7* | *60.2* | *64.5* | 62.7 | 56.6 | *60* | 56.6 | **69** | *59.8* | 37.1 |
| MDRCF | | **63.9** | **63.4** | **62.3** | **64.8** | **67.7** | **61.3** | **60.9** | **59.2** | *67.6* | **61.2** | 36.6 |
| Staple | OTB 100 | 59.5 | 53.4 | 52 | 54.3 | 55 | 54 | 54.1 | 54.9 | 47.6 | 56.1 | 39.9 |
| SRDCF | | 60 | 54.2 | 55.6 | 55 | 54.5 | 58.1 | 58.8 | 53.4 | 46.1 | 57.1 | **51.4** |
| ECO-HC | | *61.4* | *58.8* | *59.6* | *60.1* | *58.9* | *60.8* | **61.9** | *55.3* | *56.5* | *62.7* | *49.9* |
| MDRCF | | **66.3** | **60.7** | **60.6** | **60.9** | **60.5** | **63.8** | *61.8* | **58.1** | **59** | **64.3** | 49.5 |
| Staple | TC 128 | 52 | 49.4 | 48.8 | 46.3 | 55 | 41.7 | 48.1 | 47 | 38 | 49 | 35.6 |
| SRDCF | | 51.7 | 45.8 | 48.6 | 48.2 | 54 | *41.9* | 45.4 | 45.8 | 39.2 | 49.6 | 37.6 |
| ECO-HC | | *54.1* | *51.5* | *52.5* | *54.2* | *55* | **45.1** | *50.2* | *50.6* | **47** | *56.5* | **49.2** |
| MDRCF | | **57.8** | **52.9** | **52.6** | **54.7** | **57.9** | **45.1** | **50.5** | **52.5** | *46.7* | **57.6** | *46.6* |

Table II
A COMPARISON WITH SOTA TRACKERS ON THE OTB2013, OTB100 AND TC128 DATASETS. THE ENTRIES IN BOLD RED DENOTE THE BEST RESULTS AND THE ONES IN ITALIC BLUE INDICATE THE SECOND BEST. 'NO' MEANS NO RESULT

| Datasets | Trackers | Staple | SRDCF | LMCF | ECO-HC | MDRCF |
|---|---|---|---|---|---|---|
| OTB 2013 | Precision | 0.782 | 0.821 | 0.839 | *0.855* | **0.871** |
| | AUC | 0.593 | 0.619 | 0.624 | *0.641* | **0.653** |
| OTB 100 | Precision | 0.784 | 0.779 | NO | *0.835* | **0.848** |
| | AUC | 0.579 | 0.593 | 0.568 | *0.629* | **0.638** |
| TC 128 | Precision | 0.667 | 0.678 | NO | *0.743* | **0.748** |
| | AUC | 0.497 | 0.495 | NO | *0.551* | **0.556** |

To further quantitatively compare the four trackers, Table III exhibits the AUC values of the four trackers on different video attributes of the three datasets. It can be seen that MDRCF has better performance on the 11 attributes of three datasets, especially in IV, OPR, SV, OCC, DEF, MB, IPR and BC.

From the quantitative analysis data of Figure 4, Table II and Table III, it can be seen that ECO-HC and MDRCF perform better, so we apply the OTB100 dataset to conduct qualitative analysis on the trackers.

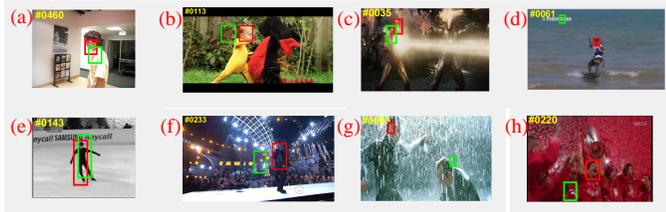

Fig.5 Qualitative comparison of ECO-HC and MDRCF on 16 sequences. Red is MDRCF, green is ECO-HC. (a) david (b) DragonBaby (c) ironman (d) KiteSurf (e) Skater (f) matrix (g) singer2 (h) soccer

The performance of the MDRCF is superior to that of the ECO-HC, which illustrates the effectiveness of the proposed tracker. Our tracker performs better on different sequences (see Figure 5), which indicates that the proposed algorithm is robust.

### D. Exponential Adaptive Merge

Based on the Staple algorithm, we test the exponential adaptive merge method proposed in this paper. Figure 6 shows the precision and AUC values of the Staple, EAMStaple and EAMStaple-PSMD on OTB2013 and OTB100, where EAMStaple-PSMD is an exponential adaptive merge method with confidence function added. Table IV shows the Frame Per Second (FPS) of Staple, EAMStaple and EAMStaple-PSMD on OTB2013 and OTB100 datasets.

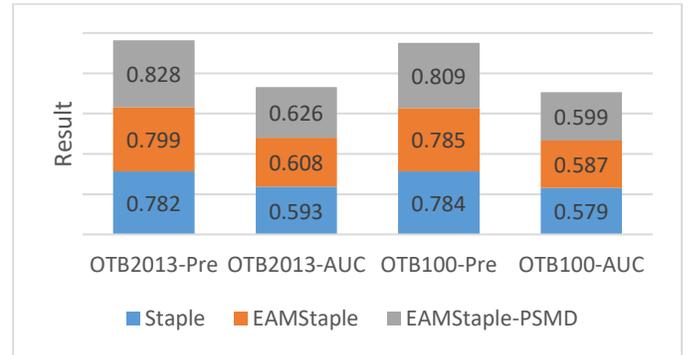

Fig.6 Tracking result of three Trackers on different datasets.

The EAMStaple presented in this paper performs better (see Figure 6), and the precision and AUC are improved on different datasets.

TABLE IV
FPS RESULTS FOR STAPLE, EAMSTAPLE AND EAMSTAPLE-PSMD ON THE OTB2013 AND OTB100 DATASETS.

| Datasets \ Trackers | Staple | EAMStaple | EAMStaple -PSMD |
|---|---|---|---|
| OTB2013 | 82.3 | *83.1* | **84.1** |
| OTB100 | 78.7 | *79.1* | **80.3** |

It can be seen from Table IV that the EAMStaple proposed in this paper doesn't cut down the FPS performance of the tracker, and even improves the FPS of the tracker to some extent.

The data in Figure 6 and Table IV also reflect that the new confidence function proposed in this paper has good robustness and better effect.

## V. CONCLUSION

In this paper, we propose a novel method for detecting the primary and secondary peaks of the characteristic response map. At the same time, a new confidence function proposed uses the adaptive update discriminant method on the discriminant mechanism to replace the traditional method of using a fixed threshold. The experimental results on multiple datasets demonstrate that both methods perform well. Furthermore, based on the proposed methods above, we propose a new robust CF tracker – MDRCF that uses hand-crafted features to improve model drift in complex scenes. Experiments indicate that MDRCF has better performances than SOTA CF trackers. Finally, we explore a simple multi-feature adaptive merge method which yield a good effect.

Although the tracker proposed in this paper performs well, there are still many needs to be improved: we solve the scale variation using the scale solution in [17], its performance is not optimal, and the follow-up work will find a solution with good performance. When searching for adaptive update discriminant thresholds, we don't use more complex functions to ensure the simplicity of the algorithm, but the cost is that the performances of the tracker are sub-optimal, so we will look for a more appropriate method in the future.